\pdfoutput=1
\documentclass[11pt,a4paper]{article}
\PassOptionsToPackage{hyphens}{url}\usepackage[draft]{hyperref}
\usepackage[hyperref]{acl2020}
\usepackage{times}
\usepackage{latexsym}
\usepackage{csquotes}
\usepackage{xspace}
\usepackage{paralist}
\usepackage{relsize}

\usepackage{microtype}

\aclfinalcopy 
\def\aclpaperid{2704}

\newcommand{\secref}[1]{Section~\ref{sec:#1}\xspace}

\newcommand\email{\begingroup \urlstyle{tt}\Url}

\title{Give Me Convenience and Give Her Death: Who Should Decide What Uses of NLP are Appropriate, and on What Basis?}

\author{Kobi Leins \qquad Jey Han Lau \qquad Timothy Baldwin \\
School of Computing and Information Systems,\\The University of
Melbourne\\
\email{{kleins,laujh,tbaldwin}@unimelb.edu.au}
}

\date{}

\begin{document}
\maketitle

\begin{abstract}
  As part of growing NLP capabilities, coupled with an awareness of the ethical dimensions of
  research, questions have been raised about whether particular datasets and tasks
   should be deemed off-limits for NLP research. We examine
  this question with respect to a paper on automatic legal sentencing
  from EMNLP 2019 which was a source of  some debate, in
  asking whether the paper should have been allowed to be published, who
  should have been charged with making such a decision, and on what basis. We focus in particular
  on the role of data statements in ethically assessing research, but
  also discuss the topic of dual use, 
  and examine
  the outcomes of similar debates in other scientific disciplines.
\end{abstract}

\section{Introduction}

NLP tools are increasingly being deployed in the wild with potentially
profound societal implications. Alongside the rise in technical
capabilities has been a growing awareness of the moral obligation of the
field to self-assess issues including: dataset and system bias
\citep{zhao-etal-2017-men}, dataset ethics
\citep{bender-friedman-2018-data}, and dual use
\citep{hovy-spruit-2016-social}. More recently, there has also been
vigorous debate on whether it is ethical for the community to work on certain topics or data types. This paper aims to
investigate this issue, focused around the examination of a paper
recently published at EMNLP 2019 on automatic prison term prediction by
\citet{chen-etal-2019-charge}. Specifically, the paper in question proposes a neural
model which performs structured prediction of the individual charges
laid against an individual, and the prison term associated with each,
which can provide an overall prediction of the prison term associated
with the case. This model was constructed using a large-scale dataset of
real-world Chinese court cases.

The primary question we attempt to address in this paper is on what
basis a given paper satisfies basic ethical requirements for
publication, in addition to examining the related question of who should
make this judgement.

Note that our intention is in no way to victimise the authors of the
paper in question, but rather to use it as a test case to objectively
ground an ethical assessment. The  authors did highlight potential 
ethical concerns of its application, but missed the point that there are 
data ethics issue in the first place. Note also that, given the topic of 
the paper, we will focus somewhat on NLP applications in the legal 
domain, but the majority of the findings/recommendations generalise and 
will be of equal relevance to other domains.


\section{Case Study in Ethical NLP Publication}

\subsection{Data ethics}

The first dimension to consider is data ethics: the data source
and procedure used to construct a dataset have an immediate impact on the
generalisabilty/interpretation of results based on that dataset, as well
as the ability for real-world harm to happen (intentionally or
otherwise) through its use. A number of proposals have recently been
made regarding documentation procedures when releasing datasets to
assist here, in particular data statements
\citep{bender-friedman-2018-data} and datasheets
\citep{Gebru+:2018}. Amalgamating the two, relevant questions to the
specific case are the following, each of which we discuss
briefly.\footnote{Note that many other important questions are covered
  in the respective frameworks, and our presentation here is biased
  towards the specific paper of interest.}

\textit{Which texts were included and what were the goals in selecting
  texts?} The dataset was constructed from published records of the
Supreme People's Court of China, following work by
\citet{DBLP:journals/corr/abs-1807-02478} in the context of a popular
shared task on automatic legal judgement prediction. The reason for
constructing this particular dataset is to ``improve the accuracy of
prison term prediction by decomposing it into a set of charge-based
prison term predictions''.

\textit{Why was the dataset created?} To enhance the structure and
granularity of earlier datasets, and achieve empirical gains in
predictive accuracy.

\textit{Were the people represented in the dataset informed about the
  data collection?} There is no mention of interaction with either
the defendants or court officials about the use of the data. The documents are in the public domain.

\textit{Was there any ethical review?}
No ethical review is mentioned in the paper.

\textit{Could this dataset expose people to harm or legal action?}  Yes, the defendants are identifiable and
the dataset directly pertains to legal action.

\textit{Does it unfairly advantage or disadvantage a particular social
  group?} The dataset does not include explicit metadata regarding the
demographics of the defendants, and the data has
first names removed, but not surnames or other named entities. It is
easy to imagine instances where the surname and location references could make the individual identifiable or
could expose demographic information, esp.\ for ethnic minorities or
areas of lower population density.

\textit{Were the people represented in the dataset provided with privacy
  guarantees?} No, no steps were taken other than removing their first names.

\textit{Does the dataset contain information that might be considered
  sensitive or confidential?} Yes, given that the labels represent
prison time served by real-world individuals, and
having personally identifying
information entombed in a dataset that potentially has longevity (cf.\
the notoriety of \textit{Pierre Vinken} from the Penn Treebank) could
potentially have direct or indirect consequences for those individuals 
and their families or group.

\textit{Does the dataset contain information that might be considered
  inappropriate or offensive?} Many of the cases are criminal in nature,
so there are potentially personal and confronting details in the court cases, including information about the victims.

\textit{How was the data annotated, and what are the demographic
  characteristics of the annotators and annotation guideline
  developers?} The ``annotation'' of the data is via court officials in
terms of their legal findings, rather than via third-party
annotations. No details are provided of the presiding court officials
and their demographics, despite there being ample evidence of
demographic bias in legal decision-making in other countries
\citep{schanzenbach2005racial,rachlinski2008does,yourstone2008evidence}.

\textit{Will the dataset be updated?} We highlight this particular question because cases can be overturned or appealed and new evidence can come to light. In this particular case,
the Supreme People's Court in China has no
legal avenue for appeal, but it is still presumably possible for a case
to be reopened on the basis of fresh evidence and a different finding
made, or overturned completely if a miscarriage of justice is found to have occurred.
On the one hand, this doesn't immediately affect the
labels in the dataset, as the sentencing is based on the facts that were
available at the time, but it could lead to situations where a legal
case which was ultimately annulled is inappropriately preserved in the
dataset in its original form, implying guilt of the individuals which
was later disproven.\\

Of these, which are relevant to whether the paper is ethically sound, or
could have made the paper less ethically questionable?  Carrying out the
research with the involvement of relevant legal authorities would
certainly have helped, in terms of incorporating domain interpretation
of the data, getting direct input as to the ultimate use of any model
trained on the data (noting that the paper does return to suggest that
the model be used in the ``Review Phase'' to help other judges
post-check judgements of presiding judges). The lack of any mention of
ethics approval is certainly troubling given the sensitivity of the
data/task. The paper does briefly mention the possibility of demographic
bias, without making any attempt to quantify or ameliorate any such
bias.  Privacy is an interesting question here, as we return to discuss
under ``data misuse'' in \secref{dual-use}, in addition to discussing
the legality of using court documents for NLP research.

Having said this, we acknowledge that similar datasets have been
constructed and used by others (esp.\
\citet{DBLP:journals/corr/abs-1807-02478}), including in major NLP
conferences (e.g.\ \citet{Zhong+:2018}, \citet{Hu+:2018}). However,
this should never be taken as a waiver for data ethic considerations. Also notable here are court proceeding datasets such as that of
\citet{Aletras+:2016}, where the use case is the prediction of the
violation of human rights (focusing on torture/degrading treatment, the
right to a fair trial, and respect for privacy), which is more clearly
aligned with ``social good''
(although there is more dataset documentation that
could have been provided in that paper, along the
lines described above).
The conversation of what social good is, though, remains an open one 
\cite{Green:2019}.

In sum, there is a level of ethical naivety and 
insensitivity in the paper, with the lack of ethics approval,
end-user engagement, and consideration of the privacy of the defendants
all being of immediate concern, but also long-term concerns including
whether NLP should be used to such ends at all.

\subsection{Dual Use}
\label{sec:dual-use}


Dual use describes the situation where a system developed for one purpose 
can be used for another.  An interesting case of dual use is OpenAI's 
GPT-2.  In February 2019, OpenAI published a technical report  
describing the development GPT-2, a very large language model that is  
trained on web data \cite{Radford+:2019}. From a science perspective, it 
demonstrates that large unsupervised language models   can be applied to a range of tasks, 
suggesting that these models have acquired some general knowledge about 
language. But another important feature of GPT-2 is its generation 
capability: it can be used to generate news articles or stories.

Due to dual-use concerns, e.g.\ fine-tuning GPT-2 to generate fake 
propaganda,\footnote{\url{https://www.middlebury.edu/institute/academics/centers-initiatives/ctec/ctec-publications-0/industrialization-terrorist-propaganda}.} 
OpenAI released only the ``small'' version of the pre-trained models. It 
was, however, not received well by the scientific community,\footnote{\url{https://thegradient.pub/openai-please-open-source-your-language-model/}.} 
with some attributing this decision to an attempt to create hype around their 
research.\footnote{\url{https://towardsdatascience.com/openais-gpt-2-the-model-the-hype-and-the-controversy-1109f4bfd5e8}.}
The backlash ultimately made OpenAI reconsidered their approach, and
release the models in stages over 9 
months.\footnote{\url{https://openai.com/blog/gpt-2-6-month-follow-up/\#fn1}.}  
During these 9 months, OpenAI engaged with other organisations to study 
the social implications of their models \cite{Solaiman+:2019}, and  
found minimal evidence of misuse, lending confidence to the publication 
of the larger models.
In November 2019 OpenAI released the  their final and largest 
model.\footnote{\url{https://openai.com/blog/gpt-2-1-5b-release/}.}

OpenAI's effort to investigate the implications of GPT-2 during the 
staged release is commendable, but this effort is voluntary, and not 
every organisation or institution will have the resources to do the 
same. It raises questions about self-regulation, and whether certain 
types of research should be pursued. A data statement is unlikely to be 
helpful here, and increasingly we are seeing more of these cases, e.g.\ 
GROVER (for generating fake news articles; \newcite{Zellers+:2019}) and 
CTRL (for controllable text generation; \newcite{Keskar+:2019}).


All of that said, for the case under consideration it is not primarily a 
question of dual use or misuse, but rather its 
\textit{primary} use: if the model
were used to inform the Supreme Court, rather than automate
decision-making, what weight should judges give the system? And what
biases has the model learned which could lead to inequities in
sentencing? It is arguable that decisions regarding human freedom, and
even potentially life and death, require greater consideration than that
afforded by an algorithm, that is, that they should not be used at all.

Although no other governments appear to be automating legal
decision-making \textit{per se}, many governments are embracing
algorithms to analyse/inform judicial decisions. In countries such as
the United States and Australia, there has been analysis of legal
decisions to understand factors such as the race/ethnicity of the defendant or the
time of the day when the judge make a decision, and how this impacts on
decision-making \cite{Zatz:Hagan:1985,Stevenson:Friedman:1994,Snowball:Weatherburn:2007,Kang+:2011}. The French government has, however, under Article 33 of
the Justice Reform Act made it illegal to analyse algorithmically any
decision made by a judge, with what some argue is the harshest possible
penalty for misconduct involving technology: a five-year
sentence.\footnote{\url{https://www.legifrance.gouv.fr/eli/loi/2019/3/23/2019-222/jo/article_33}.}

Two decades ago, Helen Nissenbaum sounded the alarm about automating 
accountability \cite{Nissenbaum:1996}. She expressed concerns that can 
be summarised in four categories. First, computerised systems are built 
by many hands and so lines of responsibility are not clear. Secondly, 
bugs are inevitable.  Third, humans like to blame the computer, which is 
problematic because of her fourth observation: that software developers 
do not like to be held responsible for their tools that they create.  Nissenbaum is not the only author who questions whether there should be limitations on certain uses of computer science \cite{Leins:2019}.

\section{Comparable Concerns in the Biological Sciences}


\blockquote{
\textit{We have consultations, which of the inventions and experiences which we have discovered shall be published, and which not; and take all an oath of secrecy for the concealing of those which we think fit to keep secret; though some of those we do reveal sometime to the State, and some not.}\\
\null\hfill Sir Francis Bacon, New Atlantis, 1626
}

The work of Ron Fouchier, a Dutch virologist, is informative in
considering publication practices in the NLP community. Fouchier discovered a way to make the bird flu H5N1 
transmissible between ferrets, and therefore potentially very harmful to 
humans. Fouchier's research extended the potential scope of the virus 
beyond its usual avian transmission routes and extended the reach of his 
research beyond his laboratory when he submitted his paper to a US 
journal. The Dutch government objected to this research being made 
public, and required Fouchier to apply for an export licence (later 
granted).  The situation raised a lot of concerns, and a lot of 
discussion at the time \cite{Enserink:2013}, as well as a series of 
national policies in 
response.\footnote{\url{https://www.jst.go.jp/crds/en/publications/CRDS-FY2012-SP-02.html}.} 
That said, Fouchier's work was not the first or last to be censored.  
Self-censorship was mentioned as early as the 17th-century by British 
philosopher Bacon, often credited with illuminating the scientific 
method \cite{Grajzl+:2019}.  Most recently, similar questions not about 
how research should be done, but whether it should be done at all, have 
arisen in the recent Chinese CRISPR-Cas 9 case, where HIV immunity in 
twins was allegedly increased, without prior ethical approval or 
oversight.\footnote{\url{https://www.technologyreview.com/s/614761/nature-jama-rejected-he-jiankui-crispr-baby-lulu-nana-paper/}.}

As the capabilities of language models and computing as a whole 
increase, so do the potential implications for  social 
disruption.
Algorithms are not likely to be transmitted virally, nor to be fatal, nor are 
they governed by export controls. Nonetheless, advances in computer 
science may present vulnerabilities of different kinds, risks of dual 
use, but also of expediting processes and embedding values that are not 
reflective of society more broadly.

\section{Who Decides Who Decides?}

Questions associated with who decides what should be published are not only legal,
as illustrated in Fouchier's work, but also fundamentally
philosophical. How should values be considered and reflected
within a community? What methodologies should be used to decide what is
acceptable and what is not? Who assesses the risk of dual use, misuse or
potential weaponisation? And who decides that potential scientific
advances are so socially or morally repugnant that they cannot be
permitted? How do we balance competing interests in light of complex
systems \cite{Foot:1967}. Much like nuclear, chemical and biological
scientists in times past, computer scientists are increasingly being
questioned about the potential applications, and long-term impact, of
their work, and should at the very least be attuned to the issues and
trained to perform a basic ethical self-assessment.

\section{Moving Forward}

Given all of the above, what should have been the course of action for
the paper in question? It is important to note that the only mentions of
research integrity/ethics in the Call for Papers relate to author
anonymisation, dual submissions, originality, and the veracity of the
research, meaning that there was no relevant mechanism for reviewers or
PC Chairs to draw on in ruling on the ethics of this or any other
submission. A recent innovation in this direction has been the adoption
of the ACM Code of Ethics by the Association for Computational
Linguistics, and explicit requirement in the EMNLP 2020 Calls for Papers
for conformance with the code:\footnote{https://2020.emnlp.org/call-for-papers}
\begin{quote}
  Where a paper may raise ethical issues, we ask that you include in the
  paper an explicit discussion of these issues, which will be taken into
  account in the review process. We reserve the right to reject papers
  on ethical grounds, where the authors are judged to have operated
  counter to the code of ethics, or have inadequately addressed
  legitimate ethical concerns with their work
\end{quote}
This is an important first step, in providing a structure for the
Program Committee to assess a paper for ethical compliance, and
potentially reject it in cases of significant concerns. Having said
this, the ACM Code of Ethics is (deliberately) abstract in its terms,
with relevant principles which would guide an assessment of the paper in
question including: 1.2 \textit{Avoid harm}; 1.4 \textit{Be fair and
  take action not to discriminate}; 1.6 \textit{Respect privacy}; 2.6
\textit{Perform work only in areas of competence}; and 3.1
\textit{Ensure that the public good is the central concern during all
  professional computing work}. In each of these cases, the
introspection present in a clearly-articulated data statement would help
ameliorate potential concerns.

What could an ethics assessment for ACL look like? Would an ethics
statement for ACL be enough to address all concerns? As argued above, it
is not clear that ACL should attempt to position itself as ethical
gatekeeper, or has the resources to do so.  And even if ACL could do so,
and wanted to do so, the efficacy of ethics to answer complex political
and societal challenges needs to be questioned \cite{Mittelstadt:2019}.

There certainly seems to be an argument for a requirement that papers describing new
datasets are accompanied by a data statement or datasheet of some form (e.g.\ as
part of the supplementary material, to avoid concerns over this using up
valuable space in the body of the paper). This still leaves the question
of what to do with pre-existing datasets: should they all be given a
free pass; or should there be a requirement for a data statement to be
retrospectively completed?

The GDPR provides some protection for the use of data, but its scope and 
geographic reach are limited. Further, the term ``anonymised'' is often 
a misnomer as even data that is classified by governments and other 
actors as ``anonymous'' can often easily be reidentified 
\cite{Culnane+:2020}.

What about code and model releases? Should there be a requirement that 
code/model releases also be subject to scrutiny for possible misuse, 
e.g.\ via a central database/registry?
As noted above, there are certainly cases where even if there are no
potential issues with the dataset, the resulting model can potentially
be used for harm (e.g.\ GPT-2). One could consider this as part of an
extension of data statements, in requiring that all code/model releases
associated with ACL papers be accompanied with a structured risk
assessment of some description, and if risk is found to exist, some
management plan be put in place. Looking to other scientific disciplines
that have faced similar issues in the past may provide some guidance for
our future.

Finally, while we have used one particular paper as a case study
throughout this paper, our intent was in no way to name and shame the
authors, but rather to use it as a case study to explore different
ethical dimensions of research publications, and attempt to foster
much broader debate on this critical issue for NLP research.










\section{Acknowledgements}

This research was supported in part by the Australian Research Council
(DP200102519 and IC170100030). The authors would like to thank Mark
Dras, Sarvnaz Karimi, and Karin Verspoor for patiently engaging in
rambling discussions which led to this hopefully less rambling paper,
and to the anonymous reviewers for their suggestions and insights.

\bibliography{anthology,acl2020}

\begin{thebibliography}{28}
\expandafter\ifx\csname natexlab\endcsname\relax\def\natexlab#1{#1}\fi

\bibitem[{Aletras et~al.(2016)Aletras, Tsarapatsanis, Preo\c{t}iuc-Pietro, and
  Lampos}]{Aletras+:2016}
Nikolaos Aletras, Dimitrios Tsarapatsanis, Daniel Preo\c{t}iuc-Pietro, and
  Vasileios Lampos. 2016.
\newblock Predicting judicial decisions of the european court of human rights:
  a natural language processing perspective.
\newblock \emph{PeerJ Computer Science}, 2.

\bibitem[{Bender and Friedman(2018)}]{bender-friedman-2018-data}
Emily~M. Bender and Batya Friedman. 2018.
\newblock \href {https://doi.org/10.1162/tacl_a_00041} {Data statements for
  natural language processing: Toward mitigating system bias and enabling
  better science}.
\newblock \emph{Transactions of the Association for Computational Linguistics},
  6:587--604.

\bibitem[{Chen et~al.(2019)Chen, Cai, Dai, Dai, and
  Ding}]{chen-etal-2019-charge}
Huajie Chen, Deng Cai, Wei Dai, Zehui Dai, and Yadong Ding. 2019.
\newblock \href {https://doi.org/10.18653/v1/D19-1667} {Charge-based prison
  term prediction with deep gating network}.
\newblock In \emph{Proceedings of the 2019 Conference on Empirical Methods in
  Natural Language Processing and the 9th International Joint Conference on
  Natural Language Processing (EMNLP-IJCNLP)}, pages 6361--6366, Hong Kong,
  China.

\bibitem[{Culnane and Leins(2020)}]{Culnane+:2020}
Chris Culnane and Kobi Leins. 2020.
\newblock Misconceptions in privacy protection and regulation.
\newblock \emph{Law in Context}, 36.

\bibitem[{Enserink(2013)}]{Enserink:2013}
Martin Enserink. 2013.
\newblock Dutch {H5N1} ruling raises new questions.
\newblock \emph{Science}, 342(6155):178--178.

\bibitem[{Foot(1967)}]{Foot:1967}
Philippa Foot. 1967.
\newblock The problem of abortion and the doctrine of double effect.
\newblock \emph{Oxford Review}, 5:5--15.

\bibitem[{Gebru et~al.(2018)Gebru, Morgenstern, Vecchione, Vaughan, Wallach,
  {Daum\'e III}, and Crawford}]{Gebru+:2018}
Timnit Gebru, Jamie Morgenstern, Briana Vecchione, Jennifer~Wortman Vaughan,
  Hanna Wallach, Hal {Daum\'e III}, and Kate Crawford. 2018.
\newblock Datasheets for datasets.
\newblock In \emph{Proceedings of the 5th Workshop on Fairness, Accountability,
  and Transparency in Machine Learning}, Stockholm, Sweden.

\bibitem[{Grajzl and Murrell(2019)}]{Grajzl+:2019}
Peter Grajzl and Peter Murrell. 2019.
\newblock Toward understanding 17th century {English} culture: A structural
  topic model of {Francis Bacon's} ideas.
\newblock \emph{Journal of Comparative Economics}, 47:111 -- 135.

\bibitem[{Green(2019)}]{Green:2019}
Ben Green. 2019.
\newblock {``Good''} isn't good enough.
\newblock In \emph{NeurIPS Joint Workshop on AI for Social Good}, Vancouver,
  Canada.

\bibitem[{Hovy and Spruit(2016)}]{hovy-spruit-2016-social}
Dirk Hovy and Shannon~L. Spruit. 2016.
\newblock \href {https://doi.org/10.18653/v1/P16-2096} {The social impact of
  natural language processing}.
\newblock In \emph{Proceedings of the 54th Annual Meeting of the Association
  for Computational Linguistics (Volume 2: Short Papers)}, pages 591--598,
  Berlin, Germany. Association for Computational Linguistics.

\bibitem[{Hu et~al.(2018)Hu, Li, Tu, Liu, and Sun}]{Hu+:2018}
Zikun Hu, Xiang Li, Cunchao Tu, Zhiyuan Liu, and Maosong Sun. 2018.
\newblock \href {https://www.aclweb.org/anthology/C18-1041} {Few-shot charge
  prediction with discriminative legal attributes}.
\newblock In \emph{Proceedings of the 27th International Conference on
  Computational Linguistics}, pages 487--498, Santa Fe, USA.

\bibitem[{Kang et~al.(2011)Kang, Bennett, Carbado, Casey, and
  Levinson}]{Kang+:2011}
Jerry Kang, Mark Bennett, Devon Carbado, Pam Casey, and Justin Levinson. 2011.
\newblock Implicit bias in the courtroom.
\newblock \emph{UCLA L. Rev.}, 59:1124--1187.

\bibitem[{Keskar et~al.(2019)Keskar, McCann, Varshney, Xiong, and
  Socher}]{Keskar+:2019}
Nitish~Shirish Keskar, Bryan McCann, Lav Varshney, Caiming Xiong, and Richard
  Socher. 2019.
\newblock {CTRL} -- a conditional transformer language model for controllable
  generation.
\newblock \emph{arXiv preprint arXiv:1909.05858}.

\bibitem[{Leins(2019)}]{Leins:2019}
Kobi Leins. 2019.
\newblock \emph{AI for better or for worse, or AI at all?}
\newblock Future Leaders.

\bibitem[{Mittelstadt(2019)}]{Mittelstadt:2019}
Brent Mittelstadt. 2019.
\newblock Principles alone cannot guarantee ethical {AI}.
\newblock \emph{Nat Mach Intell}, 1:501--507.

\bibitem[{Nissenbaum(1996)}]{Nissenbaum:1996}
Helen Nissenbaum. 1996.
\newblock Accountability in a computerized society.
\newblock \emph{Science and Engineering Ethics}, 2:25--42.

\bibitem[{Rachlinski et~al.(2008)Rachlinski, Johnson, Wistrich, and
  Guthrie}]{rachlinski2008does}
Jeffrey~J Rachlinski, Sheri~Lynn Johnson, Andrew~J Wistrich, and Chris Guthrie.
  2008.
\newblock Does unconscious racial bias affect trial judges?
\newblock \emph{Notre Dame Law Review}, 84:1195--1246.

\bibitem[{Radford et~al.(2019)Radford, Wu, Child, Luan, Amodei, and
  Sutskever}]{Radford+:2019}
Alec Radford, Jeff Wu, Rewon Child, David Luan, Dario Amodei, and Ilya
  Sutskever. 2019.
\newblock Language models are unsupervised multitask learners.
\newblock Technical report, OpenAI.

\bibitem[{Schanzenbach(2005)}]{schanzenbach2005racial}
Max Schanzenbach. 2005.
\newblock Racial and sex disparities in prison sentences: The effect of
  district-level judicial demographics.
\newblock \emph{The Journal of Legal Studies}, 34(1):57--92.

\bibitem[{Snowball and Weatherburn(2007)}]{Snowball:Weatherburn:2007}
Lucy Snowball and Don Weatherburn. 2007.
\newblock Does racial bias in sentencing contribute to indigenous
  overrepresentation in prison?
\newblock \emph{Australian \& New Zealand Journal of Criminology},
  40(3):272--290.

\bibitem[{Solaiman et~al.(2019)Solaiman, Brundage, Clark, Askell, Herbert-Voss,
  Wu, Radford, Krueger, Kim, Kreps, McCain, Newhouse, Blazakis, McGuffie, and
  Wang}]{Solaiman+:2019}
Irene Solaiman, Miles Brundage, Jack Clark, Amanda Askell, Ariel Herbert-Voss,
  Jeff Wu, Alec Radford, Gretchen Krueger, Jong~Wook Kim, Sarah Kreps, Miles
  McCain, Alex Newhouse, Jason Blazakis, Kris McGuffie, and Jasmine Wang. 2019.
\newblock Release strategies and the social impacts of language models.
\newblock \emph{arXiv preprint arXiv:1908.09203}.

\bibitem[{Stevenson and Friedman(1994)}]{Stevenson:Friedman:1994}
Bryan~A Stevenson and Ruth~E Friedman. 1994.
\newblock Deliberate indifference: Judicial tolerance of racial bias in
  criminal justice.
\newblock \emph{Wash. \& Lee L. Rev.}, 51:509--528.

\bibitem[{Xiao et~al.(2018)Xiao, Zhong, Guo, Tu, Liu, Sun, Feng, Han, Hu, Wang,
  and Xu}]{DBLP:journals/corr/abs-1807-02478}
Chaojun Xiao, Haoxi Zhong, Zhipeng Guo, Cunchao Tu, Zhiyuan Liu, Maosong Sun,
  Yansong Feng, Xianpei Han, Zhen Hu, Heng Wang, and Jianfeng Xu. 2018.
\newblock {CAIL2018:} {A} large-scale legal dataset for judgment prediction.
\newblock \emph{CoRR}, abs/1807.02478.

\bibitem[{Yourstone et~al.(2008)Yourstone, Lindholm, Grann, and
  Svenson}]{yourstone2008evidence}
Jenny Yourstone, Torun Lindholm, Martin Grann, and Ola Svenson. 2008.
\newblock Evidence of gender bias in legal insanity evaluations: A case
  vignette study of clinicians, judges and students.
\newblock \emph{Nordic Journal of Psychiatry}, 62(4):273--278.

\bibitem[{Zatz and Hagan(1985)}]{Zatz:Hagan:1985}
Marjorie~S Zatz and John Hagan. 1985.
\newblock Crime, time, and punishment: An exploration of selection bias in
  sentencing research.
\newblock \emph{Journal of Quantitative Criminology}, 1(1):103--126.

\bibitem[{Zellers et~al.(2019)Zellers, Holtzman, Rashkin, Bisk, Farhadi,
  Roesner, and Choi}]{Zellers+:2019}
Rowan Zellers, Ari Holtzman, Hannah Rashkin, Yonatan Bisk, Ali Farhadi,
  Franziska Roesner, and Yejin Choi. 2019.
\newblock Defending against neural fake news.
\newblock In \emph{Advances in Neural Information Processing Systems 32}.

\bibitem[{Zhao et~al.(2017)Zhao, Wang, Yatskar, Ordonez, and
  Chang}]{zhao-etal-2017-men}
Jieyu Zhao, Tianlu Wang, Mark Yatskar, Vicente Ordonez, and Kai-Wei Chang.
  2017.
\newblock \href {https://doi.org/10.18653/v1/D17-1323} {Men also like shopping:
  Reducing gender bias amplification using corpus-level constraints}.
\newblock In \emph{Proceedings of the 2017 Conference on Empirical Methods in
  Natural Language Processing}, pages 2979--2989, Copenhagen, Denmark.
  Association for Computational Linguistics.

\bibitem[{Zhong et~al.(2018)Zhong, Guo, Tu, Xiao, Liu, and Sun}]{Zhong+:2018}
Haoxi Zhong, Zhipeng Guo, Cunchao Tu, Chaojun Xiao, Zhiyuan Liu, and Maosong
  Sun. 2018.
\newblock Legal judgment prediction via topological learning.
\newblock In \emph{Proceedings of the 2018 Conference on Empirical Methods in
  Natural Language Processing}, pages 3540--3549, Brussels, Belgium.

\end{thebibliography}
\bibliographystyle{acl_natbib}

\end{document}